\pdfoutput=1

\documentclass[11pt]{article}

\usepackage[]{ACL}

\usepackage{times}
\usepackage{latexsym}
\usepackage{xcolor}
\usepackage{stackengine}
\usepackage{amsmath,amssymb}
\newlength\lunderset
\newlength\rulethick
\usepackage{soul}
\usepackage{tikz}

\usepackage[T1]{fontenc}

\usepackage[utf8]{inputenc}

\usepackage{microtype}
\usepackage{graphicx}
\usepackage{amsmath}
\usepackage[linesnumbered, ruled, lined]{algorithm2e}
\usepackage{booktabs}
\usepackage{multirow}
\usepackage[normalem]{ulem}
\usepackage[symbol]{footmisc}

\usepackage{tabularx}
\newcounter{gaocomm} 
  
\definecolor{blue-violet}{rgb}{0.54, 0.17, 0.89}
\definecolor{mygreen}{rgb}{0.0, 0.5, 0.0}
\definecolor{awesome}{rgb}{1.0, 0.13, 0.32}
\definecolor{bostonuniversityred}{rgb}{0.8, 0.0, 0.0}

\title{Efficient and Interpretable Compressive Text Summarisation with Unsupervised Dual-Agent Reinforcement Learning}

\author{Peggy Tang $^{\dagger}$, Junbin Gao $^{\ddagger}$, Lei Zhang $^{\diamond}$, Zhiyong Wang $^{\dagger}$ \\[-0.1cm]
$^{\dagger}$School of Computer Science, The University of Sydney   \\[-0.2cm] $^{\diamond}$International Digital Economy Academy \\[-0.2cm] $^{\ddagger}$The University of Sydney Business School, The University of Sydney \\[-0.2cm]
\tt \{peggy.tang,junbin.gao,zhiyong.wang\}@sydney.edu.au, \\[-0.3cm] \tt leizhang@idea.edu.cn
}

\begin{document}
\maketitle
\begin{abstract}
Recently, compressive text summarisation offers a balance between the conciseness issue of extractive summarisation and the factual hallucination issue of abstractive summarisation. However, most existing compressive summarisation methods are supervised, relying on the expensive effort of creating a new training dataset with corresponding compressive summaries. In this paper, we propose an efficient and interpretable  compressive summarisation method that utilises unsupervised dual-agent reinforcement learning to optimise a summary's semantic coverage and fluency by simulating human judgment on summarisation quality. 
Our model consists of an extractor agent and a compressor agent, and both agents have a multi-head attentional pointer-based structure. The extractor agent first chooses salient sentences from a document, and then the compressor agent compresses these extracted sentences by selecting salient words to form a summary without using reference summaries to compute the summary reward. 
To our best knowledge, this is the first work on unsupervised compressive summarisation. Experimental results on three widely used datasets (e.g., Newsroom, CNN/DM, and XSum) show that our model achieves promising performance and a significant improvement on Newsroom in terms of the ROUGE metric, as well as interpretability of semantic coverage of summarisation results. 
\footnote{Our source code is publicly available for research purposes at https://github.com/peggypytang/URLComSum/}
\end{abstract}
\section{Introduction}


\begin{figure}[ht!]
\centering
\includegraphics[width=.44\textwidth]{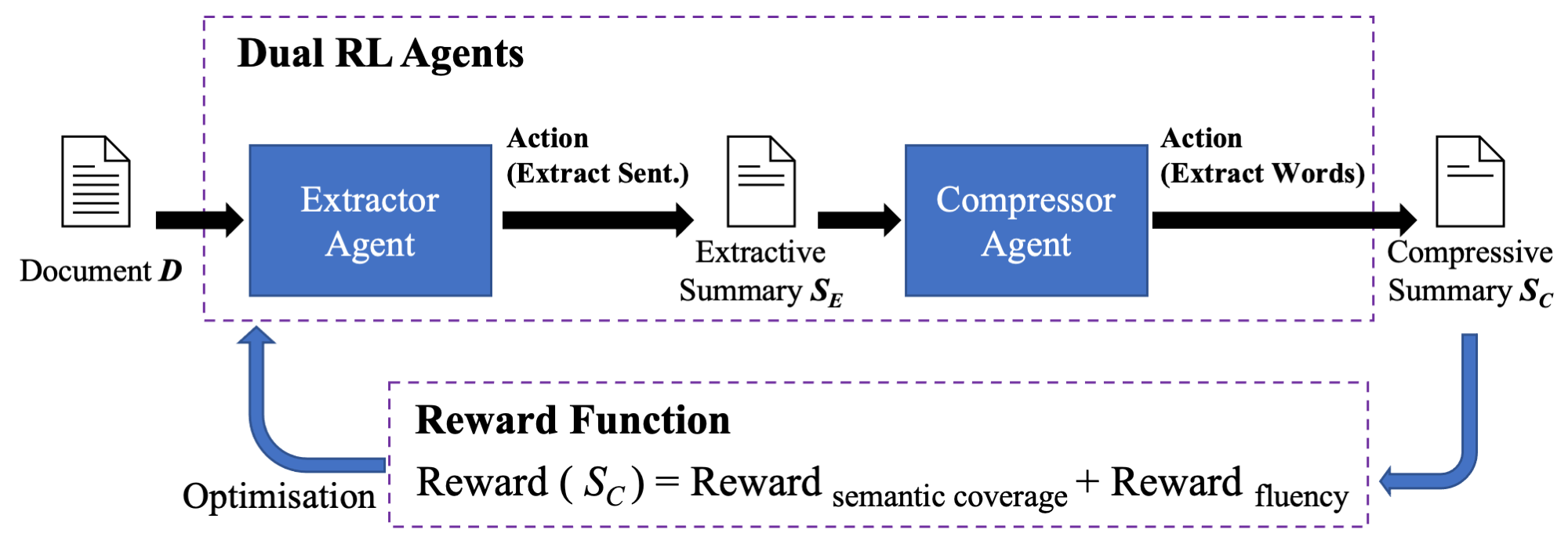}
\caption{Illustration of our proposed URLComSum. 
}
\label{fig:conceptchart}
\end{figure}

Most existing works on neural text summarisation are extractive, abstractive, and compressive-based. 
Extractive methods select salient sentences from a document to form its summary and ensure the production of grammatically and factually correct summaries. These methods usually follow the sentence ranking conceptualisation \cite{SNarayan2018Ranking,YLiu2019BERTExt,zhong2020extractive}. 
The supervised models commonly rely on creating proxy extractive training labels for training \cite{
RNallapati2017Summarunner, RJia2021TreeSum, QMa02022muchsum, SKlaus2022legal}, which can be noisy and may not be reliant. 
Various unsupervised methods \cite{HZheng2019PacSum, SXu2020unsupervised, VPadmakumar2021unsupervised,JLiu2021distanceaugmented} were proposed to leverage pre-trained language models to compute sentences similarities and select important sentences.  Although these methods have significantly improved summarisation performance, 
the redundant information that appears in the salient sentences may not be minimized effectively.


Abstractive methods formulate the task as a sequence-to-sequence generation task, with the 
document as the input sequence and the summary as the output sequence \cite{ASee2017Get, JZhang2020pegasus,LWang2021dynamicmemory,YLiu2022brio} 
As supervised learning with ground-truth summaries may not provide useful insights on human judgment approximation, reinforcement training was proposed to optimise the ROUGE metric \cite{
JParnell2021rewardsofsum}, and to fine-tune a pre-trained language model 
\cite{PLaban2020SummaryLoop}.
Prior studies showed that these generative models are highly prone to external hallucination
\cite{JMaynez2020Faithfulness}. 

Compressive summarisation is a recent appraoch which aims to select words, instead of sentences, from an input document to form a summary, which improves the factuality and conciseness of a summary. 
The formulation of compressive document summarisation is usually a two-stage extract-then-compress approach \cite{XZhang2018neural,AMendes2019Jointly,JXu2019Neural,SDesai2020Compressive}: it first extracts salient sentences from a document, then compresses the extracted sentences to form its summary. Most of these 
methods are supervised,  which require a parallel dataset with document-summary pairs to train. 
However, the ground-truth summaries of existing datasets are usually abstractive-based and do not contain supervision information needed for extractive summarisation or compressive summarisation \cite{JXu2019Neural, AMendes2019Jointly, SDesai2020Compressive}. 


Therefore, to address these limitations, we propose a novel unsupervised compressive summarisation method with dual-agent reinforcement learning strategy to mimic human judgment
, namely URLComSum. As illustrated in Figure \ref{fig:conceptchart}, URLComSum consists of two modules, an extractor agent and a compressor agent. We model the sentence and word representations using a efficient Bi-LSTM \cite{AGraves2005BiLSTM} with multi-head attention \cite{AVaswani2017attention} to capture both the long-range dependencies and the relationship between each word 
and each sentence
. We use a pointer network \cite{OVinyals2015pointer} to 
find the optimal subset of sentences and words to be extracted since the Pointer Network is well-known for tackling combinatorial optimization problems. The extractor agent uses a hierarchical multi-head attentional Bi-LSTM model for learning the sentence representation of the input document and a pointer network for extracting the salient sentences of a document given a length budget. To further compress these extracted sentences all together, the compressor agent uses a 
multi-head attentional Bi-LSTM model for learning the word representation and a pointer network for selecting the 
words to assemble a summary. 

As an unsupervised method, URLComSum does not require a parallel training dataset.
We propose an unsupervised reinforcement learning training procedure to mimic human judgment: to reward the model that achieves high summary quality in terms of semantic coverage and language fluency. Inspired by Word Mover's Distance \cite{MKusner2015WMD}, the semantic coverage reward
is measured by Wasserstein distance \cite{peyre2019computational} between the semantic distribution of the 
document and that of the summary. The fluency reward 
is measured by Syntactic Log-Odds Ratio (SLOR) \cite{APauls2012SLOR}. SLOR is a referenceless fluency evaluation metric, 
which is effective in sentence compression \cite{KKann2018FluencyEvaluation} and has better correlation to human acceptability judgments \cite{JLau2017Grammaticality}. 

The key contributions of this paper are:

\begin{itemize}
    \item We propose the first unsupervised compressive summarisation method with dual-agent reinforcement learning, namely URLComSum. 

    \item 
    We design an efficient and interpretable multi-head attentional pointer-based neural network for learning the representation and for extracting salient sentences and words. 

    \item We propose to mimic human judgment by optimising summary quality in terms of the semantic coverage reward, measured by Wasserstein distance, and the fluency reward, measured by Syntactic Log-Odds Ratio (SLOR). 

    \item Comprehensive experimental results on three widely used datasets, including CNN/DM, XSum, Newsroom, demonstrate that URLComSum achieves great performance.
\end{itemize}

\section{Related Work}

Most of the existing works on neural text summarisation are extractive, abstractive, and compressive-based. 

\subsection{Extractive Methods} 

Extractive methods usually follow the sentence ranking conceptualisation, and an encoder-decoder scheme is generally adopted. An encoder formulates document or sentence representations, and a decoder predicts extraction classification labels. 
The supervised models commonly rely on creating proxy extractive training labels for training \cite{JCheng2016Neural, RNallapati2017Summarunner, RJia2021TreeSum}, which can be noisy and may not be reliant. Some methods were proposed to tackle this issue by training with reinforcement learning \cite{SNarayan2018Ranking, LLuo2019Reading} to optimise the ROUGE metric directly. Various unsupervised methods \cite{HZheng2019PacSum, SXu2020unsupervised, VPadmakumar2021unsupervised} were also proposed to leverage pre-trained language models to compute sentences similarities and select important sentences.  Although these methods have significantly improved summarisation performance, 
since the entire sentences are extracted individually, the redundant information that appears in the salient sentences may not be minimized effectively.

\subsection{Abstractive Methods} 

Abstractive methods formulate text summarisation as a sequence-to-sequence generation task, with the source document as the input sequence and the summary as the output sequence. Most existing methods follow the supervised RNN-based encoder-decoder framework \cite{ASee2017Get,JZhang2020pegasus,LWang2021dynamicmemory,YLiu2022brio}. As supervised learning with ground-truth summaries may not provide useful insights on human judgment approximation, reinforcement training was proposed to optimise the ROUGE metric \cite{RPaulus2018DeepReinforced, JParnell2021rewardsofsum}, and to fine-tune a pre-trained language model 
\cite{PLaban2020SummaryLoop}. These models naturally learn to integrate knowledge from the training data while generating an abstractive summary. 
Prior studies showed that these generative models are highly prone to external hallucination, thus may generate contents that are unfaithful 
to the original document \cite{JMaynez2020Faithfulness}. 

\subsection{Compressive Methods}


Compressive methods select words from a given document to assemble a summary. Due to the lack of training dataset, not until recently there have emerged works for compressive summarisation \cite{XZhang2018neural,AMendes2019Jointly,JXu2019Neural,SDesai2020Compressive}. The formulation of compressive document summarisation is usually a two-stage extract-then-compress approach: it first extracts salient sentences from a document, then compresses the extracted sentences to form its summary. Most of these methods are supervised,  which require a parallel dataset with document-summary pairs to train. However, the ground-truth summaries of existing datasets are usually abstractive-based and do not contain supervision information needed for extractive summarisation or compressive summarisation.
Several reinforcement learning based methods \cite{XZhang2018neural} use existing abstractive-based datasets for training, which is not aligned for compression. Note that existing compressors often perform compression sentence by sentence. As a result, the duplicated information among multiple sentences could be overlooked. Therefore, to address these limitations, we propose a novel unsupervised compressive method by exploring the dual-agent reinforcement learning strategy to mimic human judgment and perform text compression instead of sentence compression. 

\section{Methodology}

As shown in Figure \ref{fig:conceptchart}, our proposed compressive summarisation method, namely URLComSum, consists of two components, an extractor agent and a compressor agent. 
Specifically, the extractor agent selects salient sentences from a document \(\mathbf{D}\) to form an extractive summary \(\mathbf{S_{E}}\), and then the compressor agent compresses \(\mathbf{S_{E}}\) by selecting words to assemble a compressive summary \(\mathbf{S_{C}}\). 

\subsection{Extractor Agent}

\begin{figure}[ht!]
\centering
\includegraphics[width=.45\textwidth]{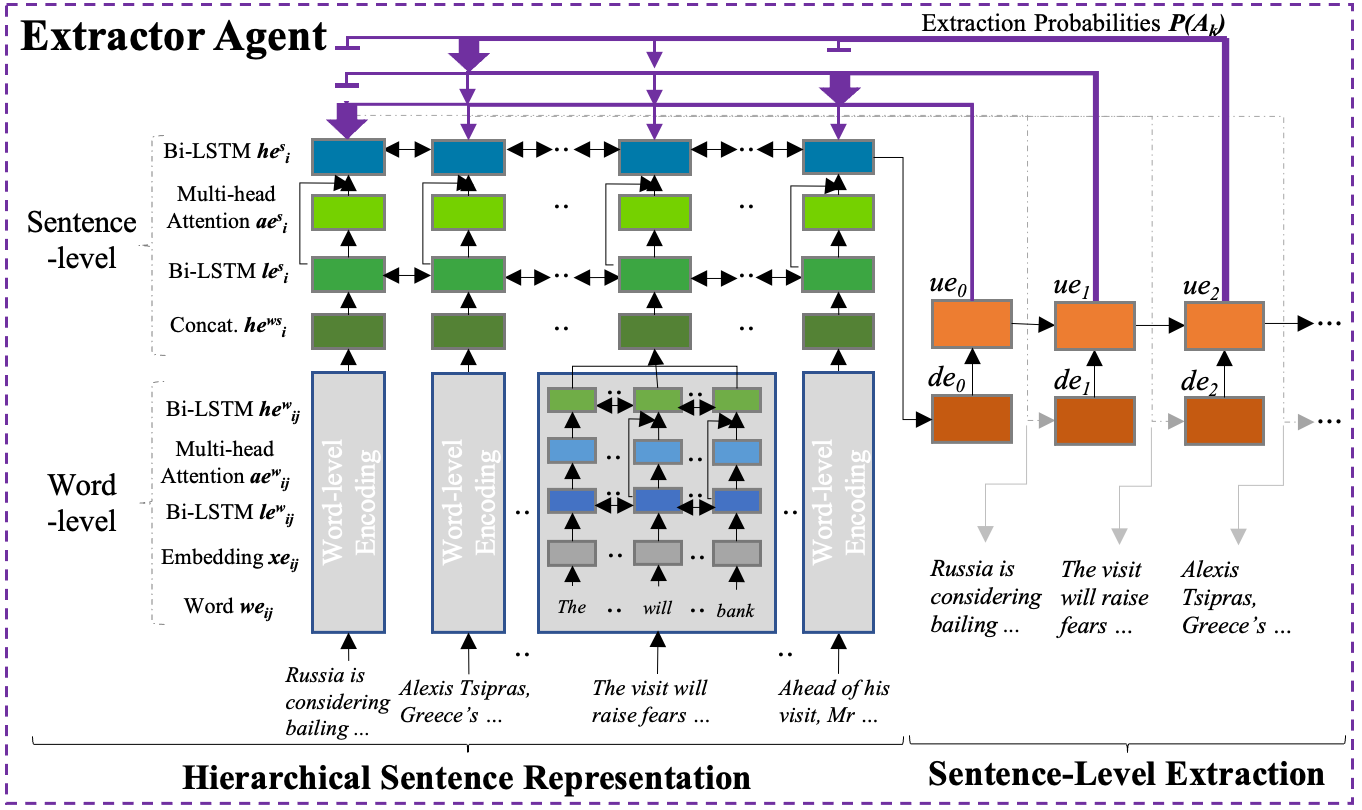}
\caption{Illustration of the extractor agent.
}
\label{fig:extractorchart}
\end{figure}

Given a document $\mathbf{D}$ consisting of a sequence of $M$ sentences $\{\mathbf{s}_{i}|i=1,...,M\}$, and each sentence $\mathbf{s}_{i}$ consisting of a sequence of $N$  words $\{\mathbf{we}_{ij}|j=1,...,N\}$\footnote{We have pre-fixed the length of each sentence and each document by padding.}, 
the extractor agent aims to produce an extractive summary \(\mathbf{S_{E}}\) by learning sentence representation and selecting \(L_{E}\) sentences from $\mathbf{D}$. 
As illustrated in Figure \ref{fig:extractorchart}, we design a hierarchical multi-head attentional 
sequential model for learning the sentence representations of the document and using a Pointer Network to extract sentences based on their representations.

\subsubsection{Hierarchical Sentence Representation}

To model the local context of each sentence 
and the global context between sentences, we use two-levels Bi-LSTMs to model this hierarchical structure, one at the word level to encode the word sequence of each sentence, one at the sentence level to encode the sentence sequence of the document. To model the context-dependency of the importance of words and sentences, we apply two levels of multi-head attention mechanism \cite{AVaswani2017attention}, one at each of the two-level Bi-LSTMs. 

Given a sentence \(\mathbf{s}_{i}\), 
we encode its words into word embeddings \(\mathbf{xe}_{i}=\{\mathbf{xe}_{ij}|j=1,...,N\}\) by \(\mathbf{xe}_{ij} = Enc(\mathbf{we}_{ij})\), where $Enc()$ denotes a word embedding lookup table. Then the sequence of word embeddings are fed into the word-level Bi-LSTM to produce an output representation of the words \(\mathbf{le}^w\):
\begin{equation}
\mathbf{le}_{ij}^w = \overleftrightarrow{\textup{LSTM}}(\mathbf{xe}_{ij}), j\in \left[ 1, N \right] .
\end{equation}
To utilize the multi-head attention mechanism to obtain \(\mathbf{ae}_i^w= \left \{ \mathbf{ae}_{i1}^w,..., \mathbf{ae}_{iN}^w \right \}\) at  word level, we define \(Q_i = \mathbf{le}_i^w\), \(K_i=V_i=\mathbf{xe}_i\), 
\begin{equation}
\mathbf{ae}_i^w  = \textup{MultiHead}(Q_i,K_i,V_i)\;.
\end{equation}
The concatenation of $\mathbf{le}_i^w$ and $\mathbf{ae}_i^w$ of the words are fed into a Bi-LSTM and the output 
is concatenated to obtain the local context representation $\mathbf{he}_i^{ws}$ for each sentence $\mathbf{s_i}$:
\begin{equation}
\begin{aligned}
\mathbf{he}_{ij}^w = \overleftrightarrow{\textup{LSTM}}(\left[ \mathbf{le}_{ij}^w ; \mathbf{ae}_{ij}^w  \right]) , j\in \left[ 1, N \right] \;, \\
\mathbf{he}_i^{ws} = \left[ \mathbf{he}_{i1}^w,..., \mathbf{he}_{iN}^w \right]\; .
\end{aligned}
\end{equation}

To further model the global context between sentences, we apply a similar structure at sentence level. \(\mathbf{he}^{ws}=\{\mathbf{he}_{i}^{ws}|i=1,...,M\}\) are fed into the sentence-level Bi-LSTM to produce output representation of the sentences \(\mathbf{le}^s\):
\begin{equation}
\mathbf{le}_i^s = \overleftrightarrow{\textup{LSTM}}(\mathbf{he}_i^{ws}), i\in \left[ 1, M \right]\; .
\end{equation}
To utilize the multi-head attention mechanism to obtain \(\mathbf{ae}^s= \left \{ \mathbf{ae}_{1}^s,..., \mathbf{ae}_{M}^s \right \}\) at sentence level, we define \(Q = \mathbf{le}^s\), \(K=V=\mathbf{he}^{ws}\),
\begin{equation}
\begin{aligned}
\mathbf{ae}^s  = \textup{MultiHead}(Q,K,V).
\end{aligned}
\end{equation}
The concatenation of the Bi-LSTM output $\mathbf{le}^s$ and the multi-head attention output $\mathbf{ae}^s$ of the sentences are fed into a Bi-LSTM to obtain the final representations of sentences \(\mathbf{he}^s= \left \{ \mathbf{he}_{1}^s,..., \mathbf{he}_{M}^s \right \}\):
\begin{equation}
\begin{aligned}
\mathbf{he}_{i}^s = \overleftrightarrow{\textup{LSTM}}(\left[ \mathbf{le}_{i}^s ; \mathbf{ae}_{i}^s  \right]) , i\in \left[ 1, M \right].
\end{aligned}
\end{equation}

\subsubsection{Sentence-Level Extraction}

Similar to \cite{YChen2018Fast}, we use an LSTM-based Pointer Network to decode the above sentence representations \(\mathbf{he}^s= \left \{ \mathbf{he}_{1}^s,..., \mathbf{he}_{M}^s \right \}\) and extract sentences recurrently to form an extractive summary \(\mathbf{S_E} = \{A_1,...,A_k,...,A_{L_E}\}\) with $L_E$ sentences, where $A_k$ denotes the $k$-th sentence extracted.


At the $k$-th time step, the pointer network receives the sentence representation of the previous extracted sentence and has hidden state ${de}_k$. It first obtains a context vector ${de}'_k$ by attending to $\mathbf{he}^s$:
\begin{equation}
\begin{aligned}
\mathbf{ue}^k_i &= v^T \tanh (W_1\mathbf{he}^s_i +W_2 de_k), i\in(1,...,M)\; , \\
\mathbf{ae}^k_i &= \textup{softmax}(\mathbf{ue}^k_i), i \in (1,...,M)\; , \\
&{de}'_k = \sum_{i=1}^{M}\mathbf{ae}^k_i \mathbf{he}^s_i \; ,
\end{aligned}
\end{equation}
where $v,W_1,W_2$ are learnable parameters of the pointer network. Then it predicts the extraction probability \(p(A_k)\) of a sentence:

\begin{equation}
\begin{aligned}
&de_k \leftarrow \left[de_k, {de}'_k\right] \;, \\
&\mathbf{ue}^k_i = v^T \tanh(W_1\mathbf{he}^s_i +W_2de_k), i \in(1,...,M) \;, \\
&p(A_k|A_1, . . . , A_{k-1}) = \textup{softmax}(\mathbf{ue}^k) \;.
\end{aligned}
\end{equation}
Decoding iterates until \(L_E\) sentences are selected to form 
\(S_E\).

\begin{figure}[h!]
\centering
\includegraphics[width=.45\textwidth]{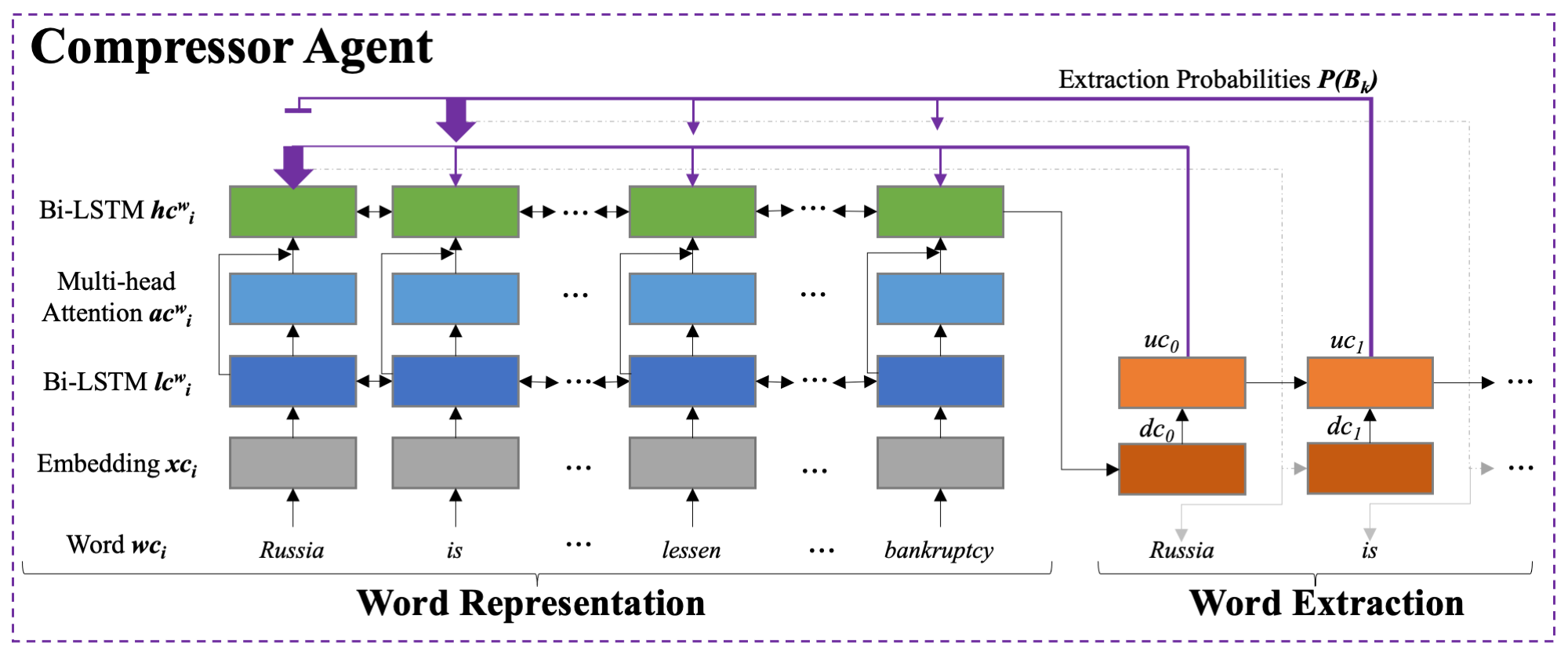}
\caption{Illustration of the compressor agent.
}
\label{fig:compressorchart}
\end{figure}

\subsection{Compressor Agent}

Given an extractive summary $\mathbf{S_E}$ consisting of a sequence of words $\mathbf{wc}=\{\mathbf{wc}_{i}|i=1,...,N\}$, the compressor agent aims to produce a compressive summary \(\mathbf{S_{C}}\) by selecting \(L_{C}\) words from $\mathbf{S_E}$. As illustrated in Figure \ref{fig:compressorchart}, it has a 
multi-head attentional Bi-LSTM model to learn the word representations. It uses a pointer network to extract words based on their representations.

\subsubsection{Word Representation}


Given a sequence of words \(\mathbf{wc}\), we encode the words into word embeddings \(\mathbf{xc}=\{\mathbf{xc}_{i}|i=1,...,N\}\) by \(\mathbf{xc}_{i} = Enc(\mathbf{wc}_{i})\). 
Then the sequence of word embeddings are fed into a Bi-LSTM to produce the words' output representation \(\mathbf{lc}^w\): 
\begin{equation}
\mathbf{lc}_{i}^w = \overleftrightarrow{\textup{LSTM}}(\mathbf{xc}_{i}), i\in \left[ 1, N \right].
\end{equation}
To utilise the multi-head attention mechanism to obtain \(\mathbf{ac}^w= \left \{ \mathbf{ac}_{1}^w,..., \mathbf{ac}_{N}^w \right \}\), we define \(Q = \mathbf{lc}^w\), \(K=V=\mathbf{xc}\),
\begin{equation}
\begin{aligned}
\mathbf{ac}^w  = \textup{MultiHead}(Q,K,V).
\end{aligned}
\end{equation}
The concatenation of $\mathbf{lc}^w$ and $\mathbf{ac}^w$ of the words are fed into a Bi-LSTM to obtain the representation $\mathbf{hc}_i^{w}$ for each word $\mathbf{wc_i}$:
\begin{equation}
\begin{aligned}
\mathbf{hc}_{i}^w = \overleftrightarrow{\textup{LSTM}}(\left[ \mathbf{lc}_{i}^w ; \mathbf{ac}_{i}^w  \right]) , i\in \left[ 1, N \right].
\end{aligned}
\end{equation}

\subsubsection{Word-Level Extraction}

The word extractor of the compressor agent shares the same structure as that of the extractor agent's sentence extractor. To select the words based on the above word representations \(\mathbf{hc}^w= \left \{ \mathbf{hc}_{1}^w,..., \mathbf{hc}_{N}^w \right \}\), the word extractor decodes and extracts words recurrently to produce \(\{B_1,...,B_k,...,B_{L_C}\}\), where $B_k$ denotes the word extracted at the $k$-th time step. 
The selected words are reordered by their locations in the input document and assembled to form the compressive summary \(\mathbf{S_C}\).

\subsection{Reward in Reinforcement Learning}

We use the compressive summary $\mathbf{S_C}$ to compute the reward of reinforcement learning and denote \(\text{Reward} (\mathbf{D}, \mathbf{S_C})\) as \(\text{Reward} (\mathbf{D}, \mathbf{S})\) for simplicity. 
\(\text{Reward} (\mathbf{D}, \mathbf{S})\) is a weighted sum of the semantic coverage award $\text{Reward}_{\text{cov}} (\mathbf{D}, \mathbf{S})$ and the fluency reward 
$ \text{Reward}_{\text{flu}}(\mathbf{S})$:
\begin{equation} \label{eqt:totalreward}
\begin{aligned}
\text{Reward} (\mathbf{D}, \mathbf{S}) = w_{\text{cov}}  \text{Reward}_{\text{cov}} (\mathbf{D}, \mathbf{S})  \\
+ w_{\text{flu}}   \text{Reward}_{\text{flu}} (\mathbf{S}) \;,
\end{aligned}
\end{equation}
where $w_{\text{cov}}$ and $w_{\text{flu}}$ denote the weights of two rewards. 

\subsubsection{Semantic Coverage Reward}

We compute 
\(\text{Reward}_{\text{cov}}\) with the Wasserstein distance between the corresponding semantic distributions of the document $\mathbf{D}$ and the summary $\mathbf{S}$, which is the minimum 
cost required to transport the semantics from $\mathbf{D}$ to $\mathbf{S}$.  
We denote $\mathbf{D}=\{d_{i}|i=1,...,N\}$ to represent a document, where $d_i$ indicates the count of the $i$-th token (i.e., word or phrase in a vocabulary of size $N$). Similarly, 
for a summary $\mathbf{S}=\{s_{j}|j=1,...,N\}$, $s_j$ is respect to the count of the $j$-th token . 
The semantic distribution of a document is characterized in terms of normalised term frequency without the stopwords.
The term frequency of the $i$-th token in the document $\mathbf{D}$ and the $j$-th token in the summary $\mathbf{S}$ are denoted as $\text{TF}_{\mathbf{D}}(i)$ and $\text{TF}_{\mathbf{S}}(j)$, respectively. 
By defining $\text{TF}_{\mathbf{D}} = \{\text{TF}_{\mathbf{D}}(i)\} \in \mathbf{R} ^{N}$ and  $\text{TF}_{\mathbf{S}} = \{\text{TF}_{\mathbf{S}}(j)\} \in \mathbf{R}^{N}$, we have the semantic distributions within 
$\mathbf{D}$ and 
$\mathbf{S}$ respectively. 

The transportation cost matrix $\mathbf{C}$ is obtained by measuring the semantic similarity between each of the tokens. Given a pre-trained tokeniser and token embedding model with $N$ tokens, define $\mathbf{v}_i$ to represent the feature embedding of the $i$-th token. Then the transport cost $c_{ij}$ from the $i$-th 
to the $j$-th token is computed based on the cosine similarity:
\(c_{ij} = 1- \frac{<\mathbf{v}_{i}, \mathbf{v}_{j}>}{\left \| \mathbf{v}_{i} \right \| _{2}\left \|  \mathbf{v}_{j} \right \|_{2} }\).
An optimal transport plan $\mathbf{T}^{*}=\{t^{*}_{i,j}\}\in\mathbf{R}^{N\times N}$ in pursuit of minimizing the transportation cost can be obtained by solving the optimal transportation and resources allocation optimization problem \cite{peyre2019computational}. 
Note that the transport plan can be used to interpret the transportation of tokens from document to summary, which brings interpretability to our URLComSum method.

Wasserstein distance measuring the distance between the two semantic distributions $\text{TF}_{\mathbf{D}}$ and $\text{TF}_{\mathbf{S}}$ with the optimal transport plan 
is computed by:
\(d_W(\text{TF}_{\mathbf{D}},\text{TF}_{\mathbf{S}}|\mathbf{C}) = \sum_{i,j} t^{*}_{ij}c_{ij} \).
\(\text{Reward}_{\text{cov}}(\mathbf{D}, \mathbf{S})\) can be further defined as:
\begin{equation} \label{eqt:coverage}
\text{Reward}_{\text{cov}}(\mathbf{D}, \mathbf{S}) = 1-d_W(\text{TF}_{\mathbf{D}},\text{TF}_{\mathbf{S}}|\mathbf{C})\;.
\end{equation}

\subsubsection{Fluency Reward}

We utilise Syntactic Log-Odds Ratio (SLOR) \cite{APauls2012SLOR} to measure 
\(\text{Reward}_{\text{flu}} (S)\), 
which is defined as: 
\(
\text{Reward}_{\text{flu}} (S)= \frac{1}{\left | S \right |}(\textup{log}(P_{LM}(S))-\textup{log}(P_{U}(S)))\;,
\)
where \(P_{LM}(S)\) denotes the probability of the summary assigned by a pre-trained language model \(LM\), \(p_{U}(S) = \prod_{t\in S}P(t)\) denotes the unigram probability for rare word adjustment, and \(|S|\) denotes the sentence length. 

We use the Self-Critical Sequence Training (SCST) method \cite{SRennie2017SCST}, since this training algorithm has demonstrated promising results in text summarisation \cite{RPaulus2018DeepReinforced, PLaban2020SummaryLoop}. For a given input document, the model produces two separate output summaries: the sampled summary $\mathbf{S}^{s}$, obtained by sampling the next pointer $t_i$ from the probability distribution at each time step $i$, and the baseline summary $\hat{\mathbf{S}}$, obtained by always picking the most likely next pointer $t$ at each $i$. The training objective is to minimise the following loss: 
\begin{equation}
\begin{aligned}
Loss = -(\text{Reward}(\mathbf{D}, \mathbf{S}^{s} )- \text{Reward}(\mathbf{D},\hat{\mathbf{S}}) ) \\  \cdot \frac{1}{N}\sum_{i=1}^{N}\textup{log} \, p(t^{s}_{i}|t^{s}_{1},... ,t^{s}_{i-1},\mathbf{D})\;,
\end{aligned}
\end{equation}
where $N$ denotes the length of the pointer sequence, which is the number of extracted sentences for the extractor agent and the number of extracted words for the compressor agent. 

Minimising the loss is equivalent to maximising the conditional likelihood of $\mathbf{S}^{s}$ if the sampled summary $\mathbf{S}^{s}$ outperforms the baseline summary $\hat{\mathbf{S}}$, i.e. $\text{Reward}(\mathbf{D}, \mathbf{S}^{s}) - \text{Reward}(\mathbf{D}, \hat{\mathbf{S}}) > 0$, thus increasing the expected reward of the model.

\section{Experiments}

\subsection{Experimental Settings}

We conducted comprehensive experiments on three widely used datasets: \textit{Newsroom} \cite{MGrusky2018newsroom}, \textit{CNN/DailyMail (CNN/DM)} \cite{KHermann2015Teaching}, and \textit{XSum} \cite{SNarayan2018XSum}. 
We set the LSTM hidden size to 150 and the number of recurrent layers to 3. 
We performed hyperparameter searching for \(w_{\text{cov}}\) 
and \(w_{\text{flu}}\) 
and decided to set \(w_{\text{cov}}= 1\) , \(w_{\text{flu}}= 2\) in all our experiments since it provides more balanced results across the datasets. We trained the URLComSum with AdamW \cite{ILoshchilov2018AdamW} with learning rate 0.01 with a batch size of 3. 
We obtained the word embedding from the pre-trained GloVe 
\cite{JPennington2014glove}. We used BERT for the pre-trained 
embedding models used for computing semantic coverage reward
. 
We chose GPT2 for the trained language model used for computing the fluency reward due to strong representation capacity. 

As shown in Table \ref{tab:dataset}, we followed \cite{AMendes2019Jointly} to set \(\mathbf{L_{E}}\) for Newsroom and \cite{zhong2020extractive} to set \(\mathbf{L_{E}}\) for CNN/DM and XSum. We also followed their protocols to set \(\mathbf{L_{C}}\) by matching the average number of words in summaries. 

\begin{table}[htbp]
\scriptsize
  \resizebox{.47\textwidth}{!}{
    \begin{tabular}{|l||ccc|}
    \hline
    \textbf{Dataset} & \textbf{Newsroom} & \textbf{CNN/DM} & \textbf{XSum}  \\
    \hline\hline
    \textbf{\#Sentences in Doc.} & 27  & 39    & 19      \\
    \textbf{\#Tokens in Doc.} & 659 & 766  & 367       \\
    \textbf{\(\mathbf{L_{E}}\)} & 2 & 3    & 2      \\
    \textbf{\(\mathbf{L_{C}}\)} & 26 & 58    & 24       \\
    \textbf{Train} & 995,041 & 287,113   &   204,045   \\
    \textbf{Test} & 108,862 & 11,490   &   11,334     \\
    \hline
    \end{tabular}%
    }
    \caption{Overview of the three datasets. 
    \#Sentences in Doc. and \#Tokens in Doc. denote the average number of sentences and words in the documents respectively. \(\mathbf{L_{E}}\) denotes the number of sentences to be selected by the extractor agent. \(\mathbf{L_{C}}\) denotes the number of words to be selected by the compressor agent. Train and Test denote the size of train and test sets. }
  \label{tab:dataset}%
\end{table}%

\begin{table}[h]
\footnotesize
\resizebox{.48\textwidth}{!}{
\begin{tabular}{|l||ccc|}
\hline
\textbf{Method} & ROUGE-1 & ROUGE-2 & ROUGE-L \\ \hline\hline
LEAD & 33.9 & 23.2 & 30.7 \\
LEAD-WORD & 34.9 & 23.1 & 30.7 \\\hline\hline
\multicolumn{4}{|l|}{\textbf{Supervised Methods}} \\ \hline\hline
EXCONSUMM (Ext.)* & 31.9 & 16.3 & 26.9 \\
EXCONSUMM (Ext.+Com.)* & 25.5 & 11.0 & 21.1 \\\hline\hline
\multicolumn{4}{|l|}{\textbf{Unsupervised Methods}}  \\ \hline\hline
SumLoop (Abs.)  & 27.0 & 9.6 & 26.4 \\
TextRank (Ext.)  & 24.5 & 10.1 & 20.1 \\ 
\textbf{URLComSum (Ext.)} & \ul{33.9} & \textbf{23.2} & \ul{30.0} \\ 
\textbf{URLComSum (Ext.+Com.)} & \textbf{34.6} & \ul{22.9} & \textbf{30.5} \\ 
\hline
\end{tabular}
}
\caption{Comparisons 
on the \textbf{Newsroom} test set. 
The symbol * indicates that the model is not directly comparable to ours as it is based on a subset (the "Mixed" 
) of the dataset.}
\label{tab:newsroom_rouge}
\end{table}

\begin{table}[h]
\footnotesize
\resizebox{.48\textwidth}{!}{
\begin{tabular}{|l||ccc|}
\hline
\textbf{Method} & ROUGE-1 & ROUGE-2 & ROUGE-L \\ \hline\hline
LEAD & 40.0 & 17.5 & 32.9 \\
LEAD-WORD & 39.7 & 16.6 & 32.5 \\\hline\hline
\multicolumn{4}{|l|}{\textbf{Supervised Methods}} \\ \hline\hline
LATENTCOM (Ext.)  & 41.1 & 18.8 & 37.5 \\ 
LATENTCOM (Ext.+Com.)  & 36.7 & 15.4 & 34.3 \\ 
JECS (Ext.)  & 40.7 & 18.0 & 36.8 \\ 
JECS (Ext.+Com.)  & 41.7 & 18.5 & 37.9 \\ 
EXCONSUMM (Ext.)  & 41.7 & 18.6 & 37.8 \\ 
EXCONSUMM (Ext.+Com.)  & 40.9 & 18.0 & 37.4 \\ 
CUPS (Ext.) & 43.7 & 20.6 & 40.0 \\
CUPS (Ext.+Com.) & 44.0 & 20.6 & 40.4 \\\hline\hline

\multicolumn{4}{|l|}{\textbf{Unsupervised Methods}}  \\ \hline\hline
SumLoop (Abs.)  & 37.7 & 14.8 & \textbf{34.7} \\
TextRank (Ext.)  & 34.1 & 12.8 & 22.5 \\ 
PacSum (Ext.)  & \textbf{40.3} & \textbf{17.6} & 24.9 \\ 
PMI (Ext.)  & 36.7 & 14.5 & 23.3 \\ 
\textbf{URLComSum (Ext.)} & \ul{40.0} & \ul{17.5} & \ul{32.9} \\ 
\textbf{URLComSum (Ext.+Com.)} & 39.3 & 16.0 & 32.2 \\ 
\hline
\end{tabular}
}
\caption{Comparisons 
between our URLComSum and the state-of-the-art methods 
on the \textbf{CNN/DM} test set. 
(Ext.), (Abs.), and (Com.) denote the method is extractive, abstractive, and compressive respectively.
} 
\label{tab:cnndm_rouge}
\end{table}

\begin{table}[h]
\footnotesize
\resizebox{.47\textwidth}{!}{
\begin{tabular}{|l||ccc|}
\hline
\textbf{Method} & ROUGE-1 & ROUGE-2 & ROUGE-L \\ \hline\hline
LEAD & 19.4 & 2.4 & 12.9 \\
LEAD-WORD & 18.3 & 1.9 & 12.8 \\\hline\hline
\multicolumn{4}{|l|}{\textbf{Supervised Methods}} \\ \hline\hline
CUPS (Ext.) & 24.2 & 5.0 & 18.3 \\
CUPS (Ext.+Com.) & 26.0 & 5.4 & 19.9 \\\hline\hline
\multicolumn{4}{|l|}{\textbf{Unsupervised Methods}}  \\ \hline\hline
TextRank (Ext.)  & 19.0 & \ul{3.1} & 12.6 \\ 
PacSum (Ext.)  & \textbf{19.4}      & 2.7     & 12.4  \\ 
PMI (Ext.)  & \ul{19.1} & \textbf{3.2} & 12.5 \\ 
\textbf{URLComSum (Ext.)} & \textbf{19.4} & 2.4 & \textbf{12.9} \\ 
\textbf{URLComSum (Ext.+Com.)} & 18.0 & 1.8 & \ul{12.7} \\ 
\hline
\end{tabular}
}
\caption{Comparisons 
on the \textbf{XSum} test set.
URLComSum (Ext.) denotes the extractive summary produced by our extractor agent. URLComSum (Ext.+Com.) denotes the compressive summary produced further by our compressor agent.
} 
\label{tab:xsum_rouge}
\end{table}


We compare our model with existing compressive methods which are all supervised, including \textit{LATENTCOM} \cite{XZhang2018neural}, \textit{EXCONSUMM} \cite{AMendes2019Jointly}, \textit{JECS} \cite{JXu2019Neural}, \textit{CUPS} \cite{SDesai2020Compressive}.  Since our method is unsupervised, we also compare it with unsupervised extractive and abstractive methods, including \textit{TextRank} \cite{RMihalcea2004Textrank}, \textit{PacSum} \cite{HZheng2019PacSum}, \textit{PMI} \cite{VPadmakumar2021unsupervised}, and \textit{SumLoop} \cite{PLaban2020SummaryLoop}. 
To better evaluate compressive methods, we followed a similar concept as LEAD baseline \cite{ASee2017Get} and created \textit{LEAD-WORD} baseline which extracts the first several words of a document as a summary.
The commonly used ROUGE metric \cite{CLin2004Rouge} is adopted. 

\subsection{Experimental Results} 

The experimental results of URLComSum on different datasets are shown in Table
\ref{tab:newsroom_rouge}, Table  \ref{tab:cnndm_rouge}  and Table \ref{tab:xsum_rouge} 
in terms of ROUGE-1, ROUGE-2 and ROUGE-L F-scores. (Ext.), (Abs.), and (Com.) denote that the method is extractive, abstractive, and compressive, respectively. 
Note that on the three datasets, LEAD and LEAD-WORD baseline are considered strong baselines in the literature and sometimes perform better than the state-of-the-art supervised and unsupervised models. As also discussed in \cite{ASee2017Get, VPadmakumar2021unsupervised}, it could be due to the Inverted Pyramid writing structure \cite{HPottker2003News} of news articles, in which important information is often located at the beginning of an article and a paragraph.


Our URLComSum method significantly outperforms all the unsupervised and supervised ones on Newsroom. This 
demonstrates the effectiveness of our proposed method. Note that, unlike supervised EXCONSUMM, our reward strategy contributes to performance improvement when the compressor agent is utilised. For example, in terms of ROUGE-L, EXCONSUMM(Ext.+Com.) does not outperform EXCONSUMM(Ext.), while URLComSum(Ext.+Com.) outperforms URLComSum(Ext.).
Similarly, 
our URLComSum method achieves the best performance among all the unsupervised methods on XSum, in terms of ROUGE-1 and -L. URLComSum underperforms in ROUGE-2, which may be due to the trade-off between informativeness and fluency. 
The improvement on Newsroom is greater than those on CNN/DM and XSum, which could be because the larger size of Newsroom is more helpful for training our model.

Our URLComSum method achieves comparable performance with other unsupervised methods on CNN/DM. Note that URLComSum does not explicitly take position information into account while some extractive methods take advantage of the lead bias of CNN/DM, such as PacSum and LEAD. Nevertheless, we observe that URLComSum(Ext.) achieves the same result as 
LEAD 
. Even though URLComSum is unsupervised, eventually the extractor agent learns to select the first few sentences of the documents, which follows the principle of the aforementioned Inverted Pyramid writing structure.

\subsubsection{Ablation Studies}

\textbf{Effect of Compression.} 
We observed that the extractive and compressive methods usually obtain better results than the abstractive ones in terms of ROUGE scores on CNN/DM and Newsroom, and vice versa on XSum. It may be that CNN/DM and Newsroom contain summaries that are usually more extractive, whereas XSum's summaries are highly abstractive. 
We noticed that URLComSum(Ext.+Com.) generally achieves higher ROUGE-1 and -L scores than its extractive version on Newsroom. Meanwhile, on CNN/DM and XSum, the compressive version has slightly lower ROUGE scores than the extractive version. We observe similar behaviour in the literature of compressive summarisation, which may be that the sentences of news articles have dense information and compression does not help much to further condense the content.

\textbf{Effect of Transformer.} 
Note that we investigated the popular transformer model \cite{AVaswani2017attention} in our proposed framework to replace Bi-LSTM for learning the sentence and word representations. However, we noticed the transformer-based agents do not perform as well as the Bi-LSTM-based ones while training from scratch with the same training procedure. The difficulties of training a transformer model have also been discussed in \cite{MPopel2018training, LLiu2020understanding}. Besides, the commonly used pre-trained transformer models, such as BERT \cite{JDevlin2019bert} and BART \cite{MLewis2020bart}, require high computational resources and usually use subword-based tokenizers. They are not suitable for URLComSum since our compressor agent points to words instead of subwords. Therefore, at this stage Bi-LSTM is a simpler and more efficient choice. Nevertheless, the transformer is a module that can be included in our framework and is worth further investigation in the future.

\textbf{Comparison of Extraction, Abstraction and Compression Approaches.}
We observed that the extraction and compressive approaches usually obtain better results than the abstractive in terms of ROUGE scores on CNN/DM and Newsroom, and vice versa on XSum. It may be because CNN/DM and Newsroom contain summaries that are usually more extractive, whereas XSum's summaries are highly abstractive. Since the ROUGE metric reflects lexical matching only and overlooks the linguistic quality and factuality of the summary, it is difficult to conclude the superiority of one approach over the others solely based on the ROUGE scores. Automatic linguistic quality and factuality metrics would be essential to provide further insights and more meaningful comparisons.

\subsection{Qualitative Analysis}

In Figure \ref{fig:samplesummary_cnndm}, \ref{fig:samplesummary_xsum}, \ref{fig:samplesummary_newsroom} in Appendix \ref{sec:appx_example}, summaries produced by URLComSum are shown together with the reference summaries of the sample documents in the CNN/DM, XSum, and Newsroom datasets. This demonstrates that our proposed URLComSum method is able to identify salient sentences and words and produce reasonably fluent summaries even without supervision information. 

\subsection{Interpretable Visualisation of Semantic Coverage}

\begin{figure}

\centering
\includegraphics[width=.4\textwidth]{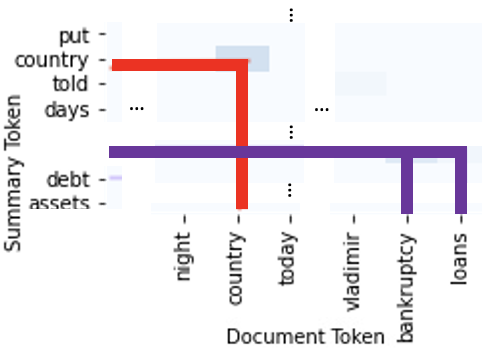}
\captionsetup{type=figure}
\caption{Interpretable visualisation of the OT plan. from a source document to a resulting summary on the CNN/DM dataset. The higher the intensity, the more the semantic content of a particular document token is covered by a summary token. \textcolor{red}{Red} line highlights the transportation from the document to the summary of semantic content of token \textcolor{red}{``country''}, which appears in both the document and the summary. \textcolor{violet}{Purple} line highlights how the semantic content of token \textcolor{violet}{``debt''}, which appears in the document only but not the summary, are transported to token \textcolor{violet}{``bankruptcy''} and \textcolor{violet}{``loans''}, which are semantically closer and have lower transport cost, and thus achieve a minimum transportation cost in the OT plan.
}
\label{fig:transport}
\end{figure}

URLComSum is able to provide an interpretable visualisation of the semantic coverage on the summarisation results through the transportation matrix. 
Figure \ref{fig:transport} illustrates the transport plan heatmap, which 
associated with a resulting summary is illustrated. A heatmap 
indicates the transportation of semantic contents 
between tokens in the document and its resulting summary. 
The higher the intensity, the more the semantic content of a particular document token is covered by a summary token. \textcolor{red}{Red} line highlights the transportation from the document to the summary of semantic content of token \textcolor{red}{``country''}, which appears in both the document and the summary. \textcolor{violet}{Purple} line highlights how the semantic content of token \textcolor{violet}{``debt''}, which appears in the document only but not the summary, are transported to token \textcolor{violet}{``bankruptcy''} and \textcolor{violet}{``loans''}, which are semantically closer and have lower transport cost, and thus achieve a minimum transportation cost in the OT plan.

\section{Conclusion}
\label{cha:conclusion}
In this paper, we have presented URLComSum, the first unsupervised and an efficient method for compressive text summarisation. Our model consists of dual agents: an extractor agent and a compressor agent. The extractor agent first chooses salient sentences from a document, and the compressor agent further select salient words from these extracted sentences to form a summary. 
To achieve unsupervised training of the extractor and compressor agents, we devise a reinforcement learning strategy to simulate human judgement on summary quality and optimize the summary's semantic coverage and fluency reward.
Comprehensive experiments on three widely used benchmark datasets demonstrate the effectiveness of our proposed URLComSum and the great potential of unsupervised compressive summarisation. Our method provides interpretability of semantic coverage of summarisation results.

\bibliography{anthology,custom}

\clearpage
\appendix
\section{Sample Summaries}
\label{sec:appx_example}

\begin{minipage}{\textwidth}

The following shows the sample summaries generated by URLComSum on the CNN/DM, XSum, and Newsroom datasets. Sentences extracted by the URLComSum extractor agent are   \hl{highlighted}. Words selected by the URLComSum compressor agent are \setulcolor{red}\ul{underlined} in red. 
Our unsupervised method URLComSum can identify salient sentences and words to produce a summary with reasonable semantic coverage and fluency. 

\centering
\includegraphics[width=.95\textwidth]{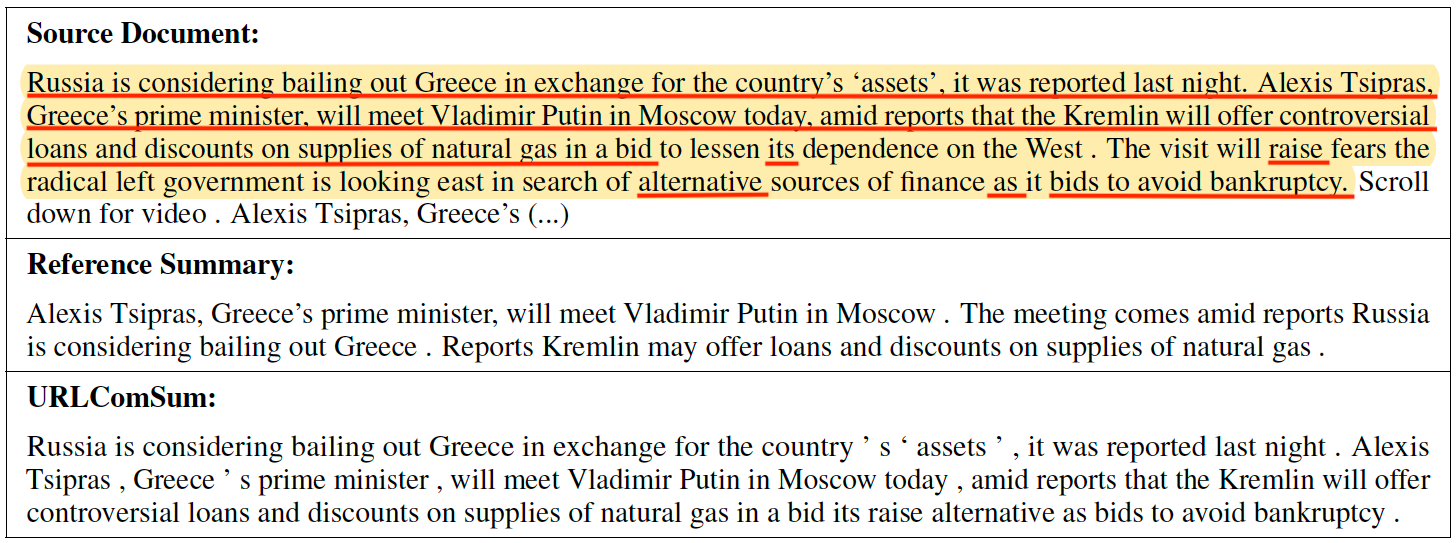}
\captionsetup{type=figure}
\caption{A sample summary produced by URLComSum on the CNN/DM dataset. The summary generated by URLComSum has ROUGE-1, ROUGE-2, and ROUGE-L F-Scores of 68.8, 52.7, and 62.4 respectively, with semantic coverage reward 0.76 and fluency reward 0.64, while the reference summary has semantic coverage reward 0.80 and fluency reward 0.62. 
}
\label{fig:samplesummary_cnndm}
\centering
\includegraphics[width=.95\textwidth]{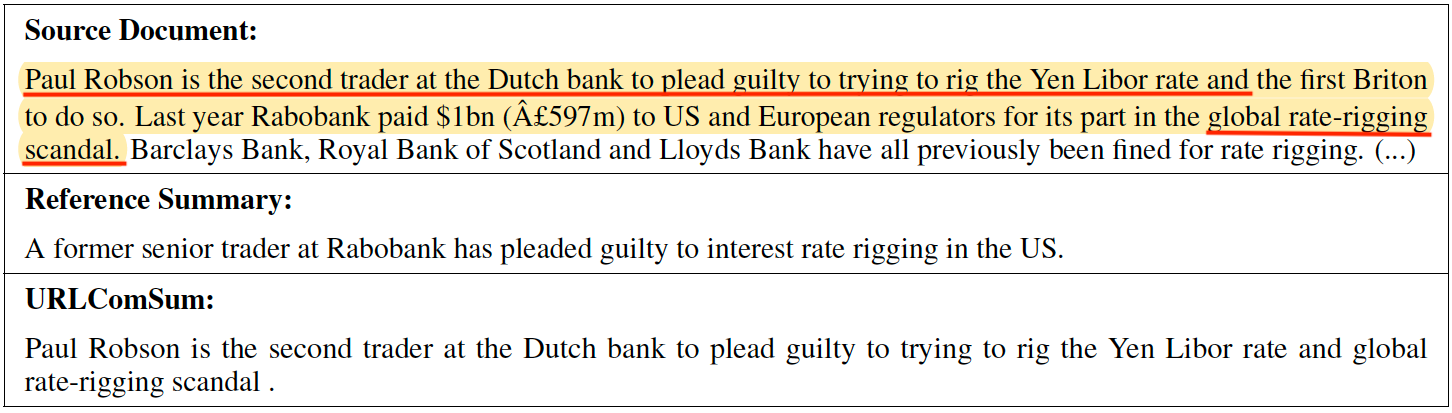}
\captionsetup{type=figure}
\caption{A sample summary produced by URLComSum on the XSum dataset. The summary generated by URLComSum has ROUGE-1, ROUGE-2, and ROUGE-L F-Scores of 38.1, 20.0, and 33.3 respectively, with semantic coverage reward 0.77 and fluency reward 0.56, while the reference summary has semantic coverage reward 0.73 and fluency reward 0.59. 
}
\label{fig:samplesummary_xsum}
\centering
\includegraphics[width=.95\textwidth]{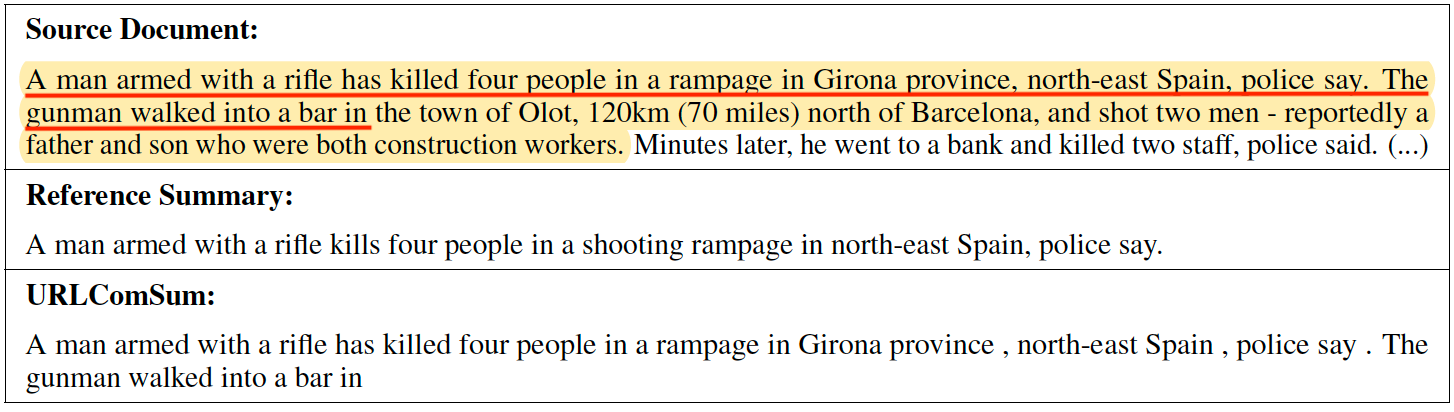}
\captionsetup{type=figure}
\caption{A sample summary produced by URLComSum on the Newsroom dataset. The summary generated by URLComSum has ROUGE-1, ROUGE-2, and ROUGE-L F-Scores of 76.6, 62.2, and 76.6 respectively, with semantic coverage reward 0.79 and fluency reward 0.61, while the reference summary has semantic coverage reward 0.76 and fluency reward 0.65. 
}
\label{fig:samplesummary_newsroom}
\end{minipage}

\clearpage

\end{document}